# Hitachi at SemEval-2020 Task 12: Offensive Language Identification with Noisy Labels using Statistical Sampling and Post-Processing

Manikandan Ravikiran[1]*, Amin Ekant Muljibhai [2]*,
Toshinori Miyoshi[2], Hiroaki Ozaki[2], Yuta Koreeda[2], Masayuki Sakata[2]
[1]Research & Development Center, Hitachi India Pvt Ltd., Bangalore, India
[2]Research & Development Group, Hitachi, Ltd., Tokyo, Japan
manikandan@hitachi.co.in
{ekant.amin.mu,toshinori.miyoshi.pd}@hitachi.com
{hiroaki.ozaki.yu,yuta.koreeda.pb,masayuki.sakata.nm}@hitachi.com

## Abstract

In this paper, we present our participation in SemEval-2020 Task-12 Subtask-A (English Language) which focuses on offensive language identification from noisy labels. To this end, we developed a hybrid system with the BERT classifier trained with tweets selected using Statistical Sampling Algorithm (SA) and Post-Processed (PP) using an offensive wordlist. Our developed system achieved $34^{th}$ position with Macro-averaged F1-score (Macro-F1) of 0.90913 over both offensive and non-offensive classes. We further show comprehensive results and error analysis to assist future research in offensive language identification with noisy labels.

## 1 Introduction

Offensive language identification focuses on identifying abusive languages, such as threats, insults, calumniation, discrimination, swearing (Zampieri et al., 2019a; Zampieri et al., 2020) which could be targeted at an individual or group. Typically, offensive language identification is helpful for (i) content moderators who face an increasing volume of abusive content and would like assistance in managing it efficiently (ii) social media sites which do auto-censoring of violence and hate speech and (iii) search query optimization (iv) improving conversations across messenger platforms (Yenala et al., 2017).

In recent times offensive language identification has seen a range of works including the creation of a large number of the corpus, open challenges, and benchmarking competitions. Moreover, with recent advances in the area of language understanding and transfer learning in NLP, offensive language identification has seen large traction (Zampieri et al., 2019b). While all these have significantly improved the capacity to identify offensive languages, there are some challenges that require addressing namely (i) Noisy labels as a result of task formulation of SemEval-2020 Task 12 Subtask-A (English Language) which focuses on the identification of offensive language with noisy labels to mitigate the problem of need of large dataset (Rosenthal et al., 2020) and (ii) identification rare and idiosyncratic offensive words (Nobata et al., 2016; Qian et al., 2018).

In this paper, we aim to address this task by developing a hybrid system with two major contributions, respectively solving previously mentioned challenges. As a first contribution, we introduce a statistical sampling algorithm that selects trustworthy tweets using statistics of noisy labels and auxiliary datasets. While sampling datasets to reduce noise has been widely explored (Frénay and Verleysen, 2014), to the best of our knowledge ours is the very first work in the context of offensive language identification. As a second contribution, we concentrate on the post-processing method that uses words from an offensive wordlist to identify tweets with rare and idiosyncratic offensive words. With these, our proposed system achieved $34^{th}$ place with Macro-averaged F1-score (Macro-F1) of 0.90913 over both offensive and non-offensive classes.

The rest of the paper is organized as follows. In section 2, we discuss literature on text classification with noisy labels. In section 3, we present the dataset of all the subtasks, followed by a detailed system description in section 4, with a focus on the statistical sampling algorithm, data splits, hyperparameters, and the post-processing approach. In section 5, we describe our results, error analysis, and some of our findings[1]. Finally, in section 6, we conclude with a summary and possible suggestions for future work.

---
*Contributed Equally

## 2 Related Work

Text classification using noisy labels in the general domain has seen works over two-decade which focuses on learning from noisy labels for a wide range of classifiers including SVM (Natarajan et al., 2013) and fisher discriminant (Lawrence and Schölkopf, 2001). While traditional predominantly approaches handle label noise by detecting and eliminating the corrupted labels, recent deep learning approaches either reduce the effect of noisy labels by using auxiliary clean dataset (Vahdat, 2017; Veit et al., 2017; Yao et al., 2019) or attempt to learn the noise distribution (Reed et al., 2014; Jiang et al., 2017; Ghosh et al., 2017). Despite a plethora of works in general domain, text classification using noisy labels is still in a nascent stage (Jindal et al., 2019; Apostolova and Kreek, 2018).

In this work, we blend the best of both traditional (Frénay and Verleysen, 2014) and deep learning approaches (Vahdat, 2017), where we introduce a simple statistical sampling algorithm, which detects trustworthy labels using the standard deviation of confidence and an auxiliary clean dataset. To the best of our knowledge, this is the first work that uses sampling datasets to handle noisy labels in the context of offensive language identification.

## 3 Dataset

The dataset for SemEval-2020 Task-12 (English Language) offensive language identification in social media that is made available to participants contains a total of 9,275,847 noisy labeled tweets (Rosenthal et al., 2020) distributed across the three subtasks (See Table 1). Subtask-A aims at the detection of offensive language (OFF or NOT). Subtask-B focuses on categorizing offensive language as targeting a specific entity (TIN) or not (UNT). Subtask-C center around identification of whether the target of an offensive post is an individual (IND), a group (GRP), or unknown (OTH). Further, each of the tweets is annotated only with two noisy labels namely $AVG_{conf}$ and $STD_{conf}$, indicating average and standard deviation for confidences generated from multiple pretrained models (Zampieri et al., 2020). While our team participated only in Subtask-A, we do use the dataset from Subtask-B for training. More details are presented in section 4.1.

|  | Subtask-A | Subtask-B | Subtask-C |
|---|---|---|---|
| **Number of tweets** | 9,087,119 | 1,88,728 | 1,13,803 |

Table 1: Training dataset statistics of SemEval 2020 Task-12 (English Language)

## 4 System Description

As mentioned earlier, for this task we developed a hybrid system with three major components namely (i) **Statistical Sampling Algorithm (SA)** that selects trustworthy tweets for training with noisy labels (ii) **Classifier** which uses a **BERT** (Devlin et al., 2019) language model fine-tuned for classification and (iii) **Post-Processing (PP)** component which uses words from wordlist to override the potentially false-negative predictions from the BERT model. Each of the three components is described in sections 4.1-4.3.

### 4.1 Statistical Sampling Algorithm (SA)

As shown in Table 1, the SemEval-2020 Task-12 Subtask-A (English Language) contains 9,087,119 tweets each with a $AVG_{conf}$ and $STD_{conf}$. While we participated only in Subtask-A, we also use the dataset from Subtask-B (1,88,728 tweets) as an auxiliary clean dataset[2] in our statistical sampling algorithm. More specifically, let $D_A$ be a dataset of Subtask-A, $D_B$ be a dataset of Subtask-B, $S_{high}$ and $S_{low}$ be upper and lower thresholds of $STD_{conf}$ to be selected, then the final dataset $D_F$ used for training is obtained by

---

[1]Disclaimer: This paper contains examples that may be considered profane, vulgar, or offensive. These contents do not reflect the views of the authors or their institutions and exclusively serve to explain linguistic research challenges.

[2]All tweets in Subtask-B dataset is offensive

1. Selecting tweets $D_F$ from $D_A$ with $STD_{conf}$= [$S_{low}$,$S_{high}$]. The intuition here is that if the majority of models used to generate the noisy labels produce similar scores, then the $STD_{conf}$ will be lower, thus we can sample tweets with the trustworthy label by using lower $STD_{conf}$. In this work, we selected tweets with [$S_{low}$,$S_{high}$]=[0.1,0.2][3] resulting in a total of 2,442,762 tweets out of 9,087,119 for training our system. The parameters $S_{high}$ and $S_{low}$ were selected through random search and validation of the developed model on the 2019 dataset.

2. Adding auxiliary clean dataset $D_B$ to $D_F$. Upon execution of step 1, we have $D_F$ with both offensive and non-offensive tweets for which we add auxiliary clean dataset $D_B$ of 1,88,728 tweets.

3. Balancing the ratio of offensive and non-offensive tweets. After step 2, the tweets may be imbalanced depending upon the $STD_{conf}$. To balance the ratio of tweets, we randomly remove the tweets from the majority class which in turn avoids bias in classification.

### 4.2 Classifier

Offensive language identification has seen extensive usage of language modeling approaches like BERT (Pelicon et al., 2019; Pavlopoulos et al., 2019; Wu et al., 2019; Liu et al., 2019), GPT (Zampieri et al., 2019b) and ELMo (Indurthi et al., 2019) with varying hyperparameters and pre-processing steps. In this work, based on its widespread usage, **BERT** (Devlin et al., 2019) is used as the classifier. In this work, we finetune the publicly available *bert-base-uncased*[4] with a sequence classification head (Wolf et al., 2019) for 10 epochs with 1k warm-up steps, batch size of 8, the learning rate of $2.0 \times 10^{-5}$, the sequence length of 64 and ADAM (Kingma and Ba, 2015) epsilon value of $1.0 \times 10^{-8}$.

### 4.3 Post-Processing (PP)

Error caused due to rare and unknown words are prevalent in offensive language identification (Nobata et al., 2016; Qian et al., 2018). More specifically, these words exhibit characteristics of low frequency due to which their meaning may not be captured by the classifier model. Hence upon obtaining predictions from the BERT model trained using samples from **SA**, in this post-processing step, we search for words from the wordlist on character-basis with subword matching process. If a word is present in the tweet we label it as OFF. In this work, we manually created a wordlist of 336 idiosyncratic and rare words by removing commonly used exclamatory offensive words from existing wordlist of Gabriel (). For example, in tweet *Glad you didn't take a pic of the **dong** I laid on his face*, we find word **dong** to be present as per our wordlist, which is an offensive slang. Thus during post-processing, we label this tweet as OFF.

## 5 Results

### 5.1 Offensive Language Identification

Table 2 and 3 show the specific performance of our developed system and a benchmark comparing the results of our developed system under the various settings respectively. Analysis of results shows various benefits and shortcomings. To begin with, we can see that the proposed approach yields absolute scores for the precision of **NOT** and recall of **OFF** classes, implying that our proposed system is useful in the identification of offensive tweets, without any false negatives (See Figure 1).

But, comparing **BERT+SA** and **BERT+SA+PP**, we can see that post-processing induces a drop in Macro-F1 of 0.05%. However, in this work, we could see that the PP gets triggered only twice, both leading to false positives (See section 5.2). Hence we conclude that the effectiveness of PP is inconclusive and is conditioned on wordlist so selected. This finding in turn warrants more testing with a different set of wordlists. Additionally, **BERT-9M** without SA and PP produces 0.2% higher Macro-F1 than our submitted system, which questions the effectiveness of SA on obtaining higher Macro-F1, despite, it can be seen that SA offers a better trade-off between trustworthy sample selection and Macro-F1, where it reduces the sample size by 7M with a mere drop of 0.2% in Macro-F1. Also, SA improves overall recall compared to **BERT-9M**, which indicates its usefulness in offensive language identification.

---

[3] It can be argued that we could select samples, just by using a single standard deviation threshold, however during our experiments having an upper and lower bound empirically helped us achieve better results.

[4] https://huggingface.co/bert-base-uncased

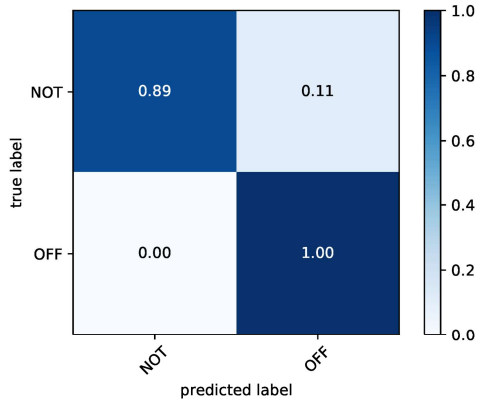

Figure 1: Confusion Matrix for our system in Subtask-A

|  | Precision | Recall | F1-Score |
|---|---|---|---|
| NOT | 1.0000 | 0.8909 | 0.9423 |
| OFF | 0.7792 | 1.0000 | 0.8759 |
| Macro Average | 0.8896 | 0.9454 | 0.9091 |

Table 2: Detailed results on 2020 testset

| Systems | Precision | Recall | F1-Score |
|---|---|---|---|
| **BERT+SA+PP (Submitted)** | 0.8896 | 0.9454 | 0.9091 |
| **BERT+SA** | 0.8901 | 0.9458 | 0.9096 |
| **BERT-9M** | 0.8933 | 0.9422 | 0.9117 |
| **BEST 2020** (Rosenthal et al., 2020) | - | - | 0.9222 |

Table 3: Benchmarking results of Subtask-A.

## 5.2 Error Analysis & Error Types

To aid future research in offensive language identification using noisy labels and to identify the cause behind 1.2% lower Macro-F1 than the top-performing system, in this section, we present detailed manual error analysis with a focus on various error types (Davidson et al., 2017; Zhang et al., 2018; van Aken et al., 2018), and their relative contributions. Since our submitted system did not produce any false negatives, we will focus on the analysis of false positives only. Also, error classes are considered irrespective of their frequency of observed samples (van Aken et al., 2018). Overall we find the following types of errors.

**Rhetorical Tweets:** Among all the error we observe rhetorical tweets to be more dominant. More specifically among the total error of 306 tweets, 168 (54%) of them are rhetorical. This is because typically while writing tweets it is a practice to use rhetorical or suggestive questions (Schmidt and Wiegand, 2017). While we don't see any targeting of tweets on individuals/groups, we do find some commonly occurring words because of which we believe our model predicted the tweets as offensive, even though they aren't. Example of those words and their occurrences[5] are ***disgusting(21), ugly(16), sick(12), sex(9), kill(9), boobs(4), fool(4)***. We think such comments could be identified using paradigmatic analysis in addition to existing signals of question words and question mark symbols.

> **Example 1:** *Damn, don't you just hate it when food is in the way of killing titans? ....*
> **Example 2:** *Donald Trump said he's a whistle-blower, .. corrupt the Democrats are... ...is he really?*

**Usage of Swear words in False Positives:** Despite learning word context, significantly classifiers always tend to focus on swear words that are often indicators of offensiveness (Zhang and Luo, 2019). So we did a manual analysis of tweets that contain swear words and not rhetorical. Examples are as shown. We find that 86 tweets (28%) show the presence of swear words and are false positives. Some of the most repeated words includes ***sucks(18), sick(15), sex(15), disgusting(6), kill(5), ugly(5), porn(3), crack(2), fuck(2), butt(2), murder(2)***. It can be noted that many of these words are used commonly in day to day speech which suggests a need for learning a better context. Further, we see that some tweets have < 5 words, thus lacking information necessary for classification.

> **Example 1:** *And today's edition of "Totally missing the fucking point" comes to you ....*
> **Example 2:** *@USER @USER @USER That sucks!*

---
[5] disgusting(21) means word disgusting appears across 21 tweets

**Humour & Irony:** Humour and Irony typically encompass swear words. Studies by Nobata et al. (2016) previously validated this in the context of hate speech detection. As such we check if the errors include humor. Again we ignore the tweets of previously described error categories. We find a total of 24 tweets (7%) is humorous and is classified as offensive by our submitted system. Further, these tweets are easily identifiable due to the presence of words such as *lol* and the presence of emoticons. Besides we do find words similar to what we described in earlier errors, which we believe is the cause for the error. Additionally, we see some tweets tend to emphasize on opposite of syntactic content.

> **Example 1:** *I gotta go downtown after this party but these fools pulled the tequila out.*
> **Example 2:** *@USER Yeah, but I suck rn lol*

**Idiosyncratic and rare words:** In section 1, we emphasized having wordlist to identify idiosyncratic tweets and tweets with rare words, owing to the issue of language change in social media. In this regard, our word list helped us identify the following two tweets as offensive. However, they turned out to be false positives causing a drop in performance.

> **Example 1:** *@USER Glad you didn't take a pic of the dong I laid on his face.*
> **Example 2:** *....Amy Ryan NAAAAILLS the accent. "this house smells like cawk*

**Doubtful labels:** Finally, we find that 6% of false-positive samples fall under the definition of offensiveness in our view, even though they are labeled otherwise. Most of them contain strong expressions directed at specific individuals. Examples are as shown. While the overall contribution to error is limited. The results do imply on presenting inter-annotator agreements for the developed dataset. Currently, no such information is available. Further, this also shows that the definition and annotation guidelines for offensive language datasets are warranted. Similar issues/suggestions were made by Waseem and Hovy (2016). Moreover, we think many of these doubtful labels are hard to annotate clearly unless there is the presence of a chain of tweets and related information surrounding the tweets.

> **Example 1:** *serges_ego you suck at sm4sh huehuehuehuehue.*
> **Example 2:** *He mess with small town women, I got bigger dreams*
> **Example 3:** *...homosexuality as normal, and homosexuality is a sin sorry if that hurts your feelings.*
> **Example 4:** *fool me twice, im killing you*

## 6 Conclusion

Offensive Language Identification is still a challenging task with multiple challenges including the need of learning from noisy labels due to lack of availability of large datasets and difficulty in identification of idiosyncratic/rare offensive words. In this work, to address these issues we developed a hybrid system with the BERT classifier trained with tweets selected using Statistical Sampling Algorithm (SA) (Section 4.1) and Post-Processing (PP) (Section 4.3) using an offensive wordlist. The Sampling Algorithm lessens the impact of the noisy label by selecting trustworthy samples, meanwhile, wordlist helps the identification of idiosyncratic and rare offensive words.

With these, our final submitted system achieved a Macro-F1 of 0.90913 in the leader board. Further, in section 5.1 we presented detailed results to find that the proposed system did not produce any false negatives. Besides, we saw that the developed system produces absolute recall for offensive tweets. At the same time, we also saw that post-processing indeed causes a drop in performance due to an increase in false positives and warranted more detailed study. Finally, in section 5.2 we detailed various errors to identify rhetorical statements, the presence of swear words as the major contributing factors, humor, and rare words are found to be minor contributors for the errors. Additionally, we also saw the issues of lack of context, lack of guidelines for offensive tweet annotation visible through labeling ambiguity for some of the tweets. In future works, in addition to addressing previously mentioned problems, we will also focus on extensive analysis of the effect of hyperparameters on results across multiple datasets.